# Comparative Analysis of Open Source Frameworks for Machine Learning with Use Case in Single-Threaded and Multi-Threaded Modes

Yuriy Kochura*[1], Sergii Stirenko[1], Anis Rojbi[2], Oleg Alienin[1], Michail Novotarskiy[1], Yuri Gordienko[1]

[1]National Technical University of Ukraine "Igor Sikorsky Kyiv Polytechnic Institute", Kiev, Ukraine
[2]University of Paris 8, Paris, France
* iuriy.kochura@gmail.com

*Abstract*—The basic features of some of the most versatile and popular open source frameworks for machine learning (TensorFlow, Deep Learning4j, and H2O) are considered and compared. Their comparative analysis was performed and conclusions were made as to the advantages and disadvantages of these platforms. The performance tests for the de facto standard MNIST data set were carried out on H2O framework for deep learning algorithms designed for CPU and GPU platforms for single-threaded and multithreaded modes of operation.

## I. INTRODUCTION

Nowadays, machine learning (ML) has advanced many fields like pedestrian detection, object recognition, visual-semantic embedding, language identification, acoustic modeling in speech recognition, video classification, etc. This success is related to the invention and application of more sophisticated machine learning models and the development of software platforms that enable the easy use of large amounts of computational resources for training such models [1]. The main aims of this paper are to review some available open source frameworks for machine learning, analyze their advantages and disadvantages, and test one of them in various computing environments including CPU and GPU-based platforms. The section *II. State of the Art* contains the short characterization of some of the most popular and versatile available open source frameworks (TensorFlow, Deep Learning4j, and H2O) for machine learning and motivation for selection of one of them for the performance tests. The section *III. Performance tests* includes description of the testing methodology, data set used, and results of these tests. The section *IV. Conclusions* dedicated to discussion of the results obtained and lessons learned.

## II. STATE OF THE ART

During the last decade numerous frameworks for machine learning appeared, but their open source implementations are seems to be most promising due to several reasons: available source codes, big community of developers and end users, and, consequently, numerous applications, which demonstrate and validate the maturity of these frameworks. Below the short characterization of the most versatile open source frameworks (Deep Learning4j, TensorFlow, and H2O) for machine learning is presented along with their comparative analysis.

### A. Deep Learning4j

Deep Learning4j (DL4J) is positioned as the open-source distributed deep-learning library written for Java and Scala that can be integrated with Hadoop and Spark [2]. It is designed to be used on distributed GPUs and CPUs platforms, and provides the ability to work with arbitrary n-dimensional arrays (also called tensors), and usage of CPU and GPU resources. Unlike many other frameworks, DL4J splits the optimization algorithm from the updater algorithm. This allows to be flexible while trying to find a combination that works best for data and problem.

### B. TensorFlow

TensorFlow is an open source software library for numerical computation was originally developed by researchers and engineers working on the Google Brain Team within Google's Machine Intelligence research organization [3] for the purposes of conducting machine learning and deep neural networks research. This software is the successor to DistBelief, which is the distributed system for training neural networks that Google has used since 2011. TensorFlow operates at large scale and in heterogeneous environments. This system uses dataflow graphs to represent computation, shared state, and the operations that mutate that state. It maps the nodes of a dataflow graph across many machines in a cluster, and within a machine across multiple computational devices, including multicore CPUs, general purpose GPUs, and custom-designed ASICs known as Tensor Processing Units (TPUs). Such architecture gives flexibility to the application developer: whereas in previous "parameter server" designs the management of shared state is built into the system, TensorFlow enables developers to experiment with novel optimizations and training algorithms.

## C. H2O

H2O software is built on Java, Python, and R with a purpose to optimize machine learning for Big Data [4]. It is offered as an open source platform with the following distinctive features. Big Data Friendly means that one can use all of their data in real-time for better predictions with H2O's fast in-memory distributed parallel processing capabilities. For production deployment a developer need not worry about the variation in the development platform and production environment. H2O models once created can be utilized and deployed like any Standard Java Object. H2O models are compiled into POJO (Plain Old Java Files) or a MOJO (Model Object Optimized) format which can easily embed in any Java environment. The beauty of H2O is that its algorithms can be utilized by various categories of end users from business analysts and statisticians (who are not familiar with programming languages using its Flow web-based GUI) to developers who know any of the widely used programming languages (e.g Java, R, Python, Spark). Using in-memory compression techniques, H2O can handle billions of data rows in-memory, even with a fairly small cluster. H2O implements almost all common machine learning algorithms, such as generalized linear modeling (linear regression, logistic regression, etc.), Naive Bayes, principal components analysis, time series, k-means clustering, Random Forest, Gradient Boosting, and Deep Learning.

## D. Comparative Analysis

From the point of view of an end user, several aspects of these frameworks are of the main interest. Except for performance and maturity, the open source frameworks could be attractive and useful, if they have the wide language and operating system support (see Table I). All of these frameworks are characterized by a quite wide ranges of supported languages and operating systems. But nowadays it is not enough in the view of the fast development of parallel and distributed computing like cluster and, especially, GPGPU computing. In this connection, TensorFlow has clear notification as to the pre-requisites for NVIDIA GPGPU cards, that should have CUDA Compute Capability (CC) 3.0 or higher. As to DL4J this is not clear because the developers stated just general support of NVIDIA GPGPU cards from GeForce GTX to Titan and Tesla that have various CC from 2.0 to 3.5. For H2O types of supported NVIDIA cards and CC are not specified, but proposed in the branching sub-framework Deep Water. The additional important aspects are the low entrance barrier and fast learning curve. They usually are based on the convenient graphical user interface, workflow management, and visualization tools. Now these features become "de facto standard" tools for integration of end users, workflows, and resources. The examples of their implementations (like WS-PGRADE/gUSE [5], KNIME [6], etc.) and applications in physics [7], chemistry [8], astronomy [9], brain-computing [10], eHealth [11] can be found elsewhere. In this context TensorFlow and H2O propose web-based graphic user interfaces TensorBoard and Flow, respectively, which are actually workflow management and visualization tools. In contrast to other frameworks H2O proposes the much shorter learning curve due to Flow, the web-based and self-explanatory user interface. In general, Flow allows end users without experience in software programming even to import remote data, create model, train it, validate it, and then save the whole workflow. In addition, the machine learning model developed in Flow can be compiled into Plain Old Java Files (POJO) format, which can be easily embedded in any Java environment. Due to these advantages, now more than 5000 organizations currently use H2O, and many well-known companies (like Cisco, eBay, PayPal, etc.) are using it for big data processing.

TABLE I.   COMPARISON OF MACHINE LEARNING FRAMEWORKS

| System (initial release) | GPU support | GUI | Operating system | Language support |
|---|---|---|---|---|
| TensorFlow (2015) | NVIDIA GPUs (CC 3.0 or higher) | TensorBoard (workflow, visualization) | Linux, macOS, Windows, Android, iOS | Python, C++ |
| DL4J (2013) | NVIDIA GPUs (Tesla, Titan) | — | Linux, macOS, Windows, Android | Java, Scala, CUDA, C, C++, Python |
| H2O (2011) | Deep Water, NVIDIA GPUs (CC not stated) | Flow (workflow, visualization, POJO) | Linux, macOS, Windows | Java, Python, R |

## III. PERFORMANCE TESTS

The performance of the mentioned frameworks was a topic of many investigations performed by developers of these frameworks and independent end users [12]. But performance of H2O was not investigated thoroughly except for its developers for unknown CPU and GPU platforms [13]. That is why H2O was selected for performance tests in this paper.

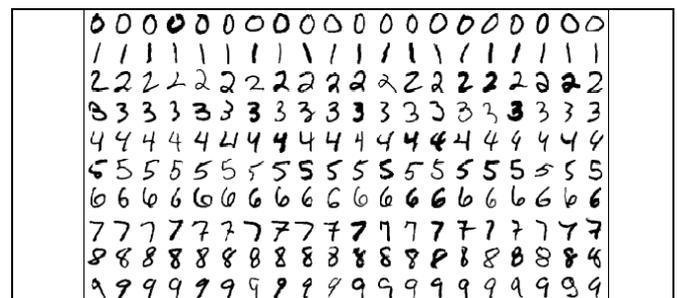
Fig. 1. The examples of the handwritten digits from MNIST data set.

The data set used in this work, called the "MNIST data," was proposed in 1998 to identify handwritten numbers. We have tested the H2O system by recognizing the handwritten digits (Fig. 1) from the publicly available MNIST data set for machine learning methods [14]. Now it is well-known "de facto standard" data set for a typical "easy-for-humans-but-hard-for-machine" problem. The used MNIST database of handwritten digits has a training set of 60,000 examples, and a test set of 10,000 examples. Each digit is represented by 28x28=784 gray-scale pixel values (features). This data set contains 785 columns. The final column is the correct answer, 0 to 9. The first 784 are the 28x28 grid of grayscale pixels, and each is 0 (for white) through to 255 (for black).

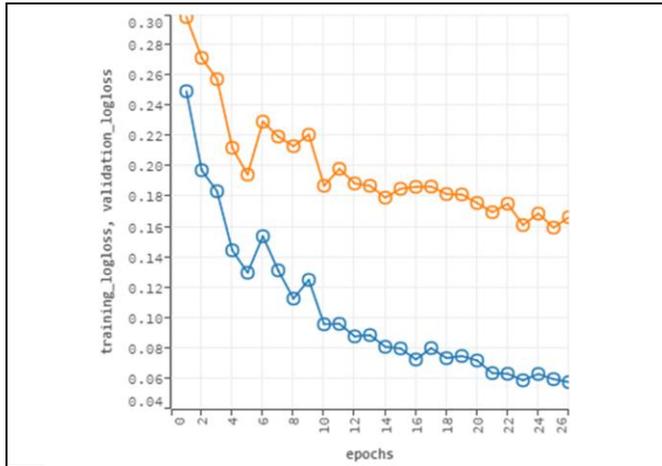

Fig. 2. Evolution of training (lower) and validation (upper) logloss values.

The tests were performed on different platforms including Intel Core i5-7200U with 4 cores (CPU1), Intel Core i7-2700K with 8 cores (CPU2), NVIDIA Tesla K40 GPU accelerator using single-threaded and multi-threaded modes of operation. The parameters of neural network were the same for the Deep Learning (CPU only) and Deep Water (CPU+GPU) algorithms. The details of these platforms and modes of operation are given below in Tables II-IV.

TABLE II. MULTI-THREADED OPERATIONS ON CPUS

| Parameter's name | Intel Core i5-7200U (CPU1) | Intel Core i7-2700K (CPU2) |
|---|---|---|
| GFLOPs | 13.85 | 29.92 |
| Duration | 2 min 18 sec | 2 min 32 sec |
| Training speed, obs/sec | 23746 | 78972 |
| Epochs | 48.5953 | 108.3821 |
| Iterations | 103 | 65 |
| Training logloss | 0.0407 | 0.0297 |
| Validation logloss | 0.1584 | 0.1616 |

The performance tests were carried out with Rectifier activation function for two algorithms Deep Learning (CPU only) and Deep Water (CPU+GPU). The stopping criterion was based on convergence of stopping_metric (equal to misclassification). The stop event occurs, if simple moving average of length $k$ of the stopping_metric does not improve for $k$:=stopping_rounds (equal to 3) scoring events. The relative tolerance for metric-based stopping criterion was equal to 0.01. The typical convergence of training (lower) and validation (upper) logloss values with epochs is shown on Fig. 2. The results of these performance tests using H2O system are presented below in Tables II-IV. It should be noted that the results of learning neural network to recognize the handwritten digits on CPUs and GPU by using multi-threaded mode of operation are inherently not reproducible due to randomization. To estimate data scattering in multi-threaded modes of operation the runs were repeated for 12 times with determination of mean and standard deviation (Table IV).

TABLE III. SINGLE-THREADED OPERATIONS ON CPUS

| Parameter's name | Intel Core i5-7200U (CPU1) | Intel Core i7-2700K (CPU2) |
|---|---|---|
| GFLOPs | 13.85 | 29.92 |
| Duration | 2 min 15 sec | 2 min 5 sec |
| Training speed, obs/sec | 13820 | 15174 |
| Epochs | 26 | 26 |
| Iterations | 26 | 26 |
| Training logloss | 0.0577 | 0.0577 |
| Validation logloss | 0.1664 | 0.1664 |

TABLE IV. MULTI-THREADED OPERATIONS ON GPU (1.43 TFLOPS)

| Parameter's name | Mean Value | Standard Deviation |
|---|---|---|
| Duration | 2 min 29sec | 17.2 sec |
| Training speed, obs/sec | 18707 | 520 |
| Epochs | 42.24 | 5.66 |
| Iterations | 2475 | 332 |
| Training logloss | 0.285 | 0.0192 |
| Validation logloss | 0.437 | 0.0236 |

IV. CONCLUSIONS

The time of convergence for logloss values with epochs was not very different for all regimes, if the standard deviation (~17 sec) of duration for multi-threaded operation on GPU will be taken into account as an estimation (Fig. 3).

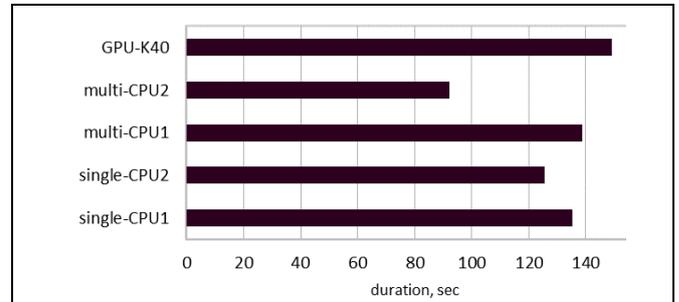

Fig. 3. Duration of training.

Despite the much higher computing power of GPU the better training speed was observed for multi-threaded regime for CPU2 with 8 cores with speedup up to 5.2 in comparison to single-threaded regime (Fig. 4). For CPU1 with 4 cores the similar speedup for multi-threaded regime was equal to 1.7 in comparison to single-threaded regime. As to GPU training speed these results can be explained by much bigger number (by ~100 times) of performed iterations.

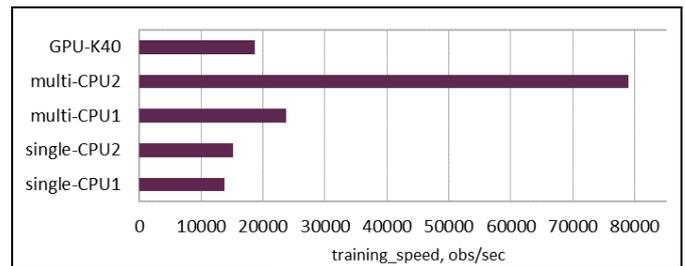

Fig. 4. Training speed.

As it is well-known the logloss values are very sensitive to outliers and this tendency is very pronounced in the case of GPU, where the much bigger iterations were used and higher training logloss values were found (Fig. 5).

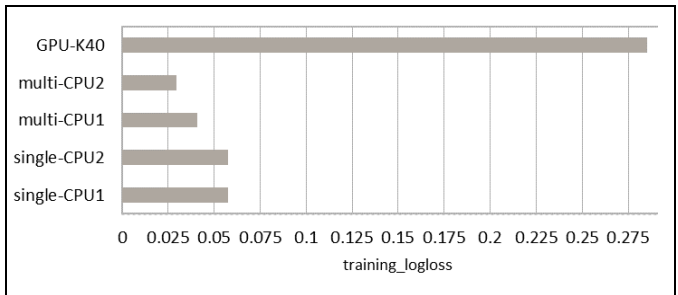

Fig. 5. Training logloss values.

The ratio of validation logloss to training logloss is equal to 1.53 for Deep Water case, which is much lower in comparison to the same ratio 2.88 for Deep Learning single-threaded case, and 3.89 and 5.44 even for Deep Learning multi-threaded case CPU1 and CPU2, respectively. This allows to make assumption that the more iterations in GPU mode give the more realistic model with the lower risk of overfitting.

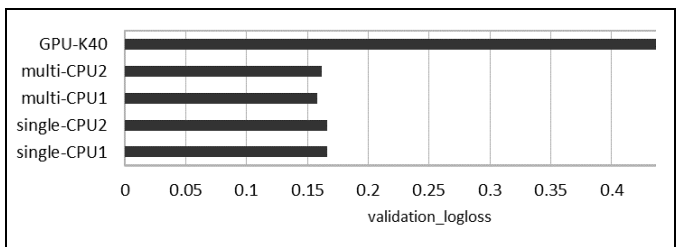

Fig. 6. Validation logloss values.

Finally, in this paper we described the basic features of some open source frameworks for machine learning, namely TensorFlow, Deep Learning4j, and H2O. For usability and performance tests H2O framework was selected. It was tested on several platforms like Intel Core i5-7200U (4 cores), Intel Core i7-2700K (8 cores), Tesla K40 GPU with the goal to evaluate their performance in the context of recognizing hand-written digits from MNIST data set. To reach this goal the same parameters of the neural network were used for Deep Learning and Deep Water algorithms. The influence of many other aspects like nature of data (for example, sparsity level and sparsity pattern), number of hidden layers and their sizes should be taken into account for the better comparative analysis, but these aspects were out of scope of the current work and will be published separately elsewhere [15,16].

The work carried out and the results obtained allow us to make the following conclusions as to H2O framework:

- H2O propose the unprecedentedly fast learning curve due to the available web-based GUI, easy workflow management tools, and visualization tools for representation of data;
- H2O allows the data scientists without any programming experience easily operate by several deep learning backends (mxnet, Caffe, TensorFlow) with various activation functions (rectifier, tahn), parameters of neural network, stopping criteria, and convergence conditions;
- H2O propose opportunities for reproducible single-threaded and non-reproducible multi-threading modes of operation for multicore CPUs and GPUs;
- multi-threaded operations on CPUs give the smaller logloss values than single-threaded operations, but the ratio of validation logloss to training logloss is much lower in comparison to multi-threaded operations on GPU, which gives the more realistic model with the lower risk of overfitting.

The work was partially supported by NVIDIA Research and Education Centers in National Technical University of Ukraine "Igor Sikorsky Kyiv Polytechnic Institute".